\def\year{2021}\relax
\documentclass[letterpaper]{article} 
\usepackage{aaai21}  
\usepackage{times}  
\usepackage{helvet} 
\usepackage{courier}  
\usepackage[hyphens]{url}  
\usepackage{graphicx} 
\def\UrlFont{\rm}  
\usepackage{graphicx}  
\usepackage{natbib}  
\usepackage{caption} 
\usepackage{amsmath}
\usepackage{amssymb}
\usepackage{booktabs}
\usepackage[ruled]{algorithm2e}
\usepackage{enumerate}
\usepackage{enumitem}

\usepackage{array}
\makeatletter
\newcommand{\thickhline}{%
    \noalign {\ifnum 0=`}\fi \hrule height 1pt
    \futurelet \reserved@a \@xhline
}
\newcolumntype{"}{@{\hskip\tabcolsep\vrule width 1pt\hskip\tabcolsep}}
\usepackage{latexsym}
\usepackage[switch]{lineno}

\begin{document}
\title{C2F-FWN: Coarse-to-Fine Flow Warping Network for \\Spatial-Temporal Consistent Motion Transfer}
\author {
        Dongxu Wei\textsuperscript{\rm 1},
        Xiaowei Xu\textsuperscript{\rm 2},
        Haibin Shen\textsuperscript{\rm 1},
        Kejie Huang\textsuperscript{\rm 1}\footnote{Corresponding author: huangkejie@zju.edu.cn}\\
}
\affiliations {
    \textsuperscript{\rm 1} Department of Information Science and Electronic Engineering, Zhejiang University \\
    \textsuperscript{\rm 2} Guangdong Provincial People's Hospital, Guangdong Academy of Medical Sciences
}
\maketitle
\begin{abstract}
Human video motion transfer (HVMT) aims to synthesize videos that one person imitates other persons' actions.
Although existing GAN-based HVMT methods have achieved great success, they either fail to preserve appearance details due to the loss of spatial consistency between synthesized and exemplary images, or generate incoherent video results due to the lack of temporal consistency among video frames.
In this paper, we propose Coarse-to-Fine Flow Warping Network (C2F-FWN) for spatial-temporal consistent HVMT. Particularly, C2F-FWN utilizes coarse-to-fine flow warping and Layout-Constrained Deformable Convolution (LC-DConv) to improve spatial consistency, and employs Flow Temporal Consistency (FTC) Loss to enhance temporal consistency.
In addition, provided with multi-source appearance inputs, C2F-FWN can support appearance attribute editing with great flexibility and efficiency.
Besides public datasets, we also collected a large-scale HVMT dataset named SoloDance for evaluation.
Extensive experiments conducted on our SoloDance dataset and the iPER dataset show that our approach outperforms state-of-art HVMT methods in terms of both spatial and temporal consistency.
Source code and the SoloDance dataset are available at https://github.com/wswdx/C2F-FWN.
\end{abstract}
\section{Introduction}
Human Video Motion Transfer (HVMT) refers to the task of synthesizing videos that one person imitates motions of other persons, which has attractive potential applications in movies, interactive games, virtual shopping, etc.
With the development of Generative Adversarial Networks (GANs) \cite{goodfellow2014generative} and GAN-based image-to-image translation techniques \cite{wang2018high,wang2018video,park2019semantic}, HVMT works have achieved great success.\par
In general, existing HVMT methods have two main streams: personalized HVMT and general-purpose HVMT. Personalized methods \cite{chan2019everybody,liu2019neural} focus on learning the mapping from motion inputs (e.g., body poses or semantic layouts that describe the desired motions) to video frames for a specific person, with a large number of frames from this person collected as the training data to fit the model for his/her appearance. To generate videos for another person, they have to perform a new round of data collection and model training, which requires massive human resources and computation costs.
The recent emergence of general-purpose methods \cite{wang2019few,liu2019liquid,wei2020gac} manages to solve this by providing additional appearance inputs (e.g., exemplary images that describe the desired appearances) for GANs. Thus they can generate videos for new persons by altering the input exemplary images.
However, most of these methods directly utilize GANs to generate values for all the pixels from scratch without preserving their spatial consistency with pixels in the exemplary images, which results in the loss of appearance details such as decorative patterns and colors of clothes.
Besides, they either don't consider temporal consistency or only focus on implicit temporal consistency among frame images when synthesizing videos, which causes low temporal coherence in their video results.
Moreover, most of them don't support HVMT with fully editable appearances, lacking flexibility and efficiency for real applications.\par
In this paper, to address these limitations, we propose Coarse-to-Fine Flow Warping Network (C2F-FWN) to ensure both spatial and temporal consistency for HVMT.
For spatial consistency, our C2F-FWN synthesizes motion transfer videos through warping based on coarse-to-fine transformation flows rather than direct generation based on GANs. Thus we can precisely model geometric deformations caused by motions to preserve spatial correlations between synthesized and exemplary image pixels.
Moreover, Layout-Constrained Deformable Convolution (LC-DConv) is utilized to extract deformable features for C2F-FWN, further improving the spatial consistency.
For temporal consistency, we propose Flow Temporal Consistency (FTC) Loss with optical flows as the constraints to enforce explicit temporal consistency among transformation flows instead of frame images, radically ensuring the video coherence.\par
\begin{figure*}[t]
\centering
\includegraphics[width=0.9\textwidth]{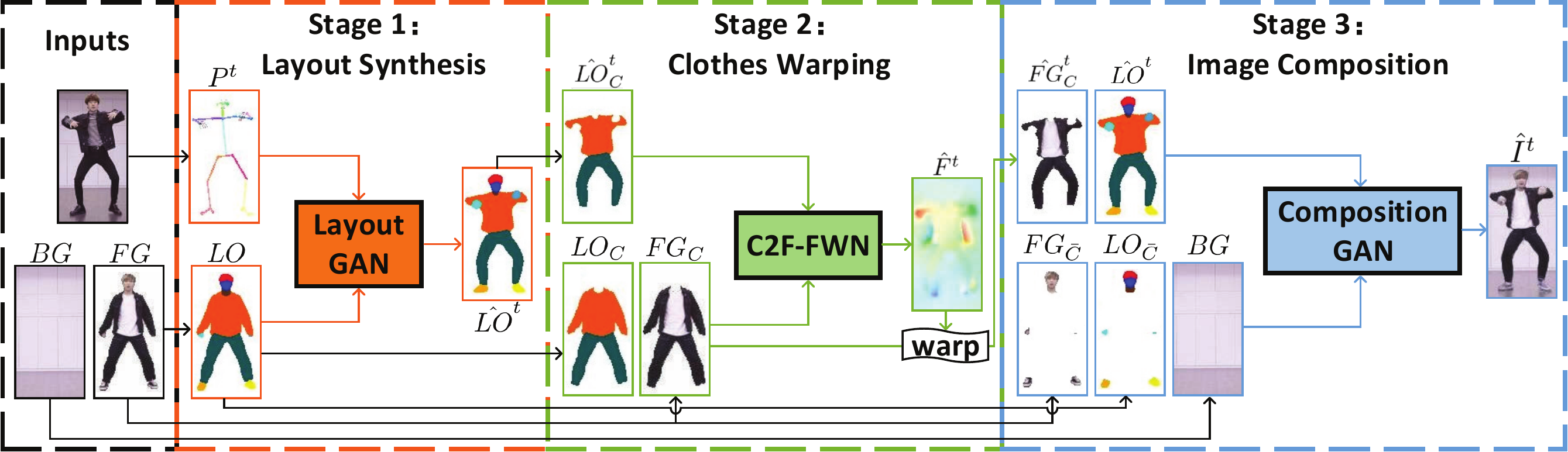} 
\caption{Overview of our method. Orange, green and blue rectangles specify processes in Stages 1, 2 and 3, respectively. Black arrows denote ordinary data flows like pose detection, layout detection and body division. Orange, green and blue arrows denote data flows of the layout GAN in Stage 1, our C2F-FWN in Stage 2, and the composition GAN in Stage 3, respectively.}
\label{overview}
\end{figure*}
In our experiments, we evaluate our method on both iPER dataset \cite{liu2019liquid} and a large-scale SoloDance dataset collected by ourselves. Both quantitative and qualitative results demonstrate that videos generated by our method have significantly better spatial and temporal consistency than existing personalized and general-purpose methods. We also show that our approach can utilize multi-source appearance inputs to enable full appearance attribute editing (e.g., change identities, tops, bottoms, backgrounds) for HVMT, which has promising application prospects.
\section{Related Work}
\subsection{Personalized HVMT}
Personalized HVMT \cite{chan2019everybody,liu2019neural,aberman2019deep,yang2020transmomo} only learns the mappings from motion inputs to video frames, with appearances learned individually in different models. Once trained, one model can only generate videos with specific appearances. To generate videos with new appearances, they need to train new models. Although such approaches can generate high-fidelity videos, they lack the efficiency for applications.
\subsection{General-Purpose HVMT}
General-purpose HVMT can be divided into direct generation methods \cite{wang2019few,wei2020gac} and warping-based methods \cite{liu2019liquid,dong2018soft,han2019clothflow}. Both utilize additional appearance inputs to control the synthesized appearances in addition to motions.
\subsubsection{Direct Generation Methods}
leverage GAN-based image-to-image translation techniques \cite{wang2018high,wang2018video,park2019semantic} to generate video frames from appearance and motion inputs directly.
\citeauthor{wang2019few} utilize several SPADE blocks \cite{park2019semantic} to adaptively propagate the appearance information throughout the network, which can achieve appearance control by altering the appearance inputs. \citeauthor{wei2020gac} propose an appearance-consistency discriminator to force the generator to generate appearances consistent with the alterable appearance inputs, which also achieves the appearance control.
However, these methods don't consider the spatial consistency between pixels in outputs and appearance inputs. Thus they can't preserve appearance details such as textures and colors well.
Moreover, they only consider temporal consistency among frame images, which is implicit and hard to learn.
On account of mode collapse and over-fitting problems of GANs \cite{webster2019detecting}, these methods often obtain low-fidelity results.
\subsubsection{Warping-Based Methods} focus on generating images through warping to preserve spatial consistency. \citeauthor{dong2018soft} utilize Thin-Plate-Spline (TPS) transformation for warping to align features of appearance inputs with those of motion inputs before GAN-based generation.
Similar feature warping can also yield fancy facial animation results for face video synthesis \cite{chen2020puppeteergan}.
Unfortunately, the TPS transformation is decided by a few control points, which restricts its warping capability due to the low degree of freedom. Thus it can't precisely model the geometric deformations. Moreover, \citeauthor{liu2019liquid} propose liquid warping based on 3D SMPL models to achieve similar feature alignment. However, the SMPL models \cite{loper2015smpl} only describe naked human bodies. Thus they can't model surface deformations for clothes and hair.
Instead of warping features, \citeauthor{han2019clothflow} propose to warp images using flows, which is similar to our approach in spirit. However, they directly estimate the dense flow field from misaligned features of appearance and motion inputs, failing to model large deformations when the two inputs greatly differ from each other in motion.
Besides, none of these warping-based methods considers the temporal consistency between warping operations of neighbored frames, making them not capable of synthesizing coherent videos.
Moreover, these methods all adopt standard convolutions in their networks, where the fixed receptive fields can't accommodate shape variances. Hence they can't extract appropriate features for human subjects with various shapes to estimate warping functions.
\section{Method}
\subsection{Overview}
The overview of our method is shown in Figure \ref{overview}, which contains three stages: layout synthesis (Stage 1), clothes warping (Stage 2) and image composition (Stage 3). For ease of discussion, the used symbols are presented as follows.
Given an exemplary foreground image $FG$ describing the desired human appearance and an exemplary background image $BG$ describing the desired background appearance, we aim at synthesizing videos that the exemplary human foreground $FG$ performs motions described by the pose sequence $P^{1\sim{T}}$ in the exemplary background $BG$.
For the exemplary foreground $FG$, we detect its semantic layout ${LO}$ and divide it to further obtain layout $LO_{C}$ and foreground $FG_{C}$ for the clothing parts (i.e., tops and bottoms), as well as layout $LO_{\bar{C}}$ and foreground $FG_{\bar{C}}$ for the non-clothing parts (i.e., hair, face, torso and shoes).
Provided with these processed inputs, we synthesize the corresponding output video sequence $\hat{I}^{1\sim{T}}$ by three stages described in Figure \ref{overview}. Since $\hat{I}^{1\sim{T}}$ is generated frame by frame, we take the synthesis of the $t$-th frame $\hat{I}^t$ as an example for brevity.\par
\textbf{In Stage 1}, we utilize a layout GAN to generate the semantic layout $\hat{LO}^t$, which has the same motion as $P^{t}$ and the same appearance as $LO$. We further divide $\hat{LO}^t$ to obtain the clothing layout $\hat{LO}_C^t$.
\textbf{In Stage 2}, taking $\chi_1=\{{LO}_{C},FG_{C}\}$ as the appearance input and taking $\chi_2=\hat{LO}_C^t$ as the motion input, our C2F-FWN computes the transformation flow $\hat{F}^t$ to warp the exemplary clothing foreground $FG_{C}$ into the foreground $\hat{FG}_C^t$, which precisely aligns with the generated clothing layout $\hat{LO}_C^t$.
\textbf{In Stage 3}, we utilize a composition GAN to generate the remaining parts including the non-clothing foreground and the background, and compose them with the clothing foreground $\hat{FG}_C^t$ from Stage 2 to generate the full frame image $\hat{I}^t$.
Note that we don't generate the non-clothing parts through warping for two reasons. First, the appearance of the non-clothing parts varies sharply in different views, making it extremely hard to model their appearance changes through warping. Second, texture and color patterns of the non-clothing parts are simple and easy to generate using GANs. Therefore, we utilize the composition GAN to synthesize the non-clothing parts.\par
Particularly, the layout GAN and the composition GAN follow the Vid2Vid design presented in \cite{wang2018video}.
Vid2Vid is a general image-to-image translation backbone consisting of two encoders and two decoders ($E_1$,$E_2$,$D_1$,$D_2$ for brevity). $E_1$ and $E_2$ aim to encode features for two inputs $\mathcal{I}_1$ (i.e., current conditional inputs) and $\mathcal{I}_2$ (i.e., previous generated results), respectively. $D_1$ and $D_2$ aim to decode the added features of $\mathcal{I}_1$ and $\mathcal{I}_2$ to output $\mathcal{O}_1$ (i.e., a raw result) and $\mathcal{O}_2$ (i.e., an optical flow). Then we can obtain the current frame result by using $\mathcal{O}_2$ to warp the last frame result and add it to $\mathcal{O}_1$.
For the \textbf{layout GAN}, $\mathcal{I}_1$ denotes the concatenated $\{P^t,LO\}$. $\mathcal{I}_2$ denotes the concatenated $\{\hat{LO}^{t-1},\hat{LO}^{t-2}\}$. Thus we can utilize the Vid2Vid backbone to generate $\hat{LO}^t$. Besides, to better synthesize the one-hot semantic layouts rather than RGB images, we replaced image reconstruction losses of Vid2Vid with a structure-sensitive pixel-wise softmax loss introduced in human parsing works \cite{liang2018look}.
Similarly, for the \textbf{composition GAN}, $\mathcal{I}_1$ denotes the concatenated $\{\hat{LO}^t,\hat{FG}_C^t,LO_{\bar{C}},FG_{\bar{C}},BG\}$. $\mathcal{I}_2$ denotes the concatenated $\{\hat{I}^{t-1},\hat{I}^{t-2}\}$. The Vid2Vid backbone learns to automatically attach non-clothing parts to clothes synthesized in Stage 2, and thus obtain the full image $\hat{I}^t$.\par
In the following, the details of our \textbf{C2F-FWN} including coarse-to-fine flow warping, Layout-Constrained Deformable Convolution (LC-DConv) and Flow Temporal Consistency (FTC) Loss are presented. At last, the unique characteristic of our C2F-FWN, multi-source appearance attribute editing is discussed.
\begin{figure}[t]
\centering
\includegraphics[width=1\columnwidth]{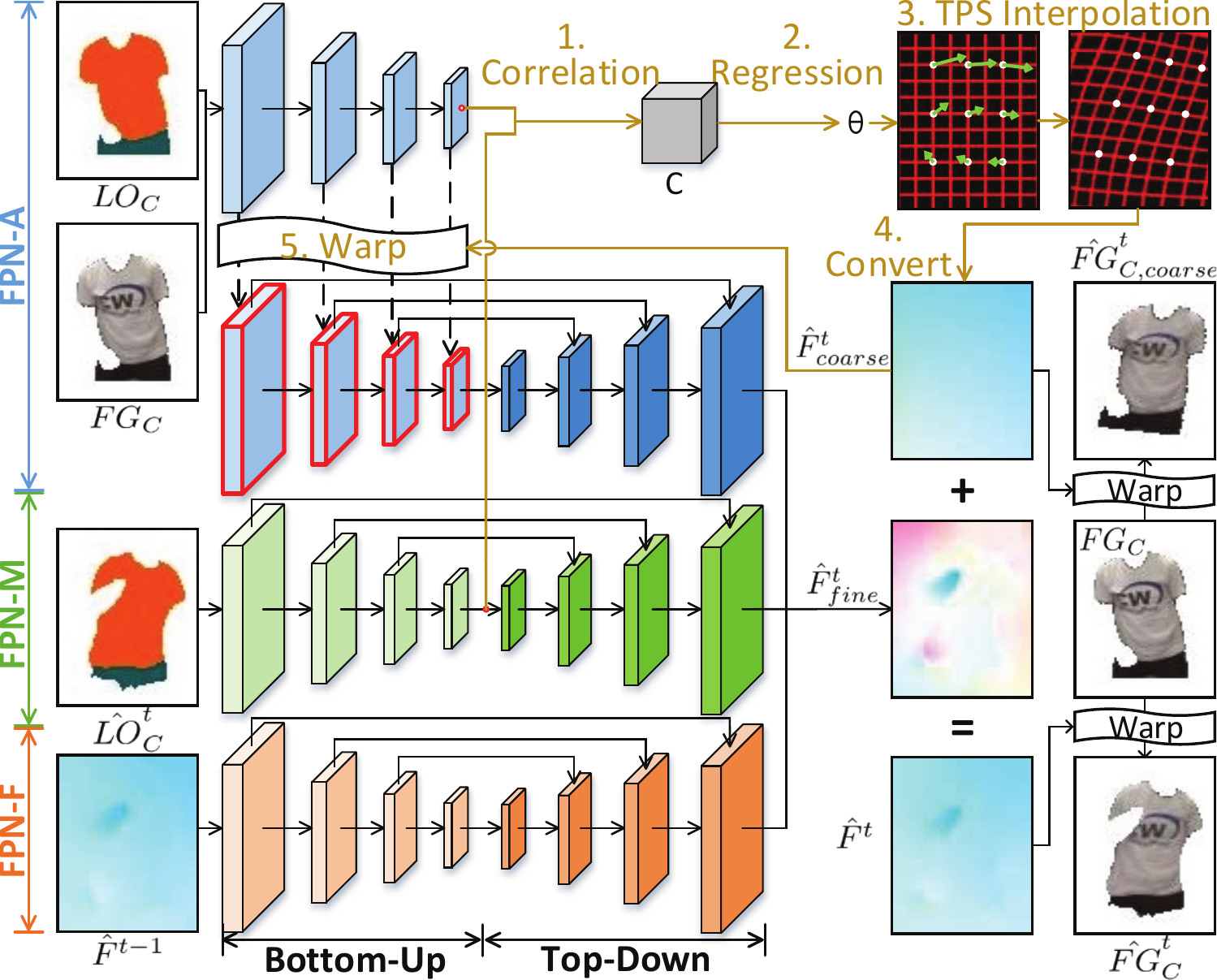} 
\caption{Illustration of C2F-FWN. We use feature maps in different colors to differentiate the three FPNs, where blue, green, and orange feature maps specify FPN-A, FPN-M, and FPN-F, respectively. Each FPN has two pathways connected by lateral connections (horizontal arrows), where we use light and dark colors to differentiate the features of bottom-up and top-down pathways, respectively. Steps 1$\sim$5 drawn in gold describe the procedure of our coarse flow warping.}
\label{stage2}
\end{figure}
\subsection{Coarse-to-Fine Flow Warping}
Before diving into details of the coarse-to-fine flow warping, we first explain its motivation and mechanism.
In cases that ${LO}_{C}$ greatly differs from $\hat{LO}_C^t$ in motion, there would be far distances between pixels in the appearance input $\chi_1$ and their correlated pixels in the motion input $\chi_2$.
If we estimate the whole transformation flow directly based on the concatenation of misaligned appearance and motion features, we would fail because standard convolutions with limited kernel sizes can't build correlations between pixels far away from each other in position.
Differently, our C2F-FWN first estimates a coarse Thin-Plate-Spline (TPS) flow $\hat{F}_{coarse}^t$ based on the smallest bottom-up features to coarsely warp the appearance features into the desired motion, where the effects of far distances w.r.t. the size of inputs can be ignored due to the large receptive fields of small-size features.
Thus the appearance and the motion features are aligned. Then we can concatenate the largest top-down appearance and motion features to further compute the refinement flow $\hat{F}_{fine}^t$ for fine warping, where the effects of far distances have been eliminated after preliminary feature alignment.\par
As shown in Figure \ref{stage2}, C2F-FWN contains three feature pyramid networks (FPN) \cite{lin2017feature}: FPN-A, FPN-M and FPN-F, responsible for extracting pyramidal features of appearance input $\chi_1=\{{LO}_{C},{FG}_{C}\}$, motion input $\chi_2=\hat{LO}_C^t$ and previously estimated transformation flow $\chi_3=\hat{F}^{t-1}$. Each FPN has two symmetrical pathways (bottom-up and top-down). Specifically, the top-down pathway is built upon the bottom-up pathway via lateral connections, with sizes of the bottom-up features growing smaller and sizes of the top-down features growing larger. Benefiting from such symmetric design, both coarse and fine warpings can be realized in the unified C2F-FWN.\par
\subsubsection{Coarse Flow Warping}
The procedure of our coarse flow warping is described in steps 1$\sim$5 in Figure \ref{stage2}.
First, we compute the correlation map $C$ with each position containing the pairwise similarities between the smallest bottom-up features of FPN-A and FPN-M.
The correlation map is then fed into a regression layer to compute $K{\times}2$ parameters ($\theta$), which represent positions of $K$ control points ($K$=$3{\times}3$ in this paper). Based on TPS interpolation \cite{rocco2017convolutional}, we can generalize the mapping between the estimated $K$ control points and their corresponding predefined grid points to all the pixels of ${FG}_{C}$, and hence move each pixel to its new position to obtain the coarsely warped clothes $\hat{FG}_{C,coarse}^t$.
To enable the supervision of the coarse warping, we utilize a VGG loss \cite{johnson2016perceptual} $L^{coarse}_{VGG}$ to minimize the difference between $\hat{FG}_{C,coarse}^t$ and the ground truth.\par
Then, to make the TPS transformation compatible with our transformation flow, we convert it to a coarse flow $\hat{F}_{coarse}^t$ by computing the position difference before and after transformation for each pixel. Let $P=(x,y)$ denote the position of a pixel in the warped clothes $\hat{FG}_{C,coarse}^t$, and let $P^{'}=(x^{'},y^{'})$ denote the position of the same pixel in the exemplary clothes ${FG}_{C}$. The coarse flow at position $(x,y)$ can be given by: $\hat{F}_{coarse}^{t}(x,y)=\overrightarrow{PP^{'}}=(x^{'}-x,y^{'}-y)$, which is the same for all the other positions.\par
Then we downsample $\hat{F}_{coarse}^t$ to different sizes to warp all the bottom-up features of FPN-A, roughly aligning them with the generated layout $\hat{LO}_C^t$, which represents the desired motion. Thus we can compute the corresponding roughly-aligned top-down appearance features via lateral connections, with pixels located at positions close to the correlated pixels in the top-down motion features, facilitating the subsequent estimation of the refinement flow.
\subsubsection{Fine Flow Warping}
As shown in Figure \ref{stage2}, we predict the refinement flow $\hat{F}_{fine}^t$ based on the concatenation of the largest top-down features of the three FPNs, where we include features of FPN-F to allow for learning the temporal consistency with previous transformation flows. Specifically, the refinement flow $\hat{F}_{fine}^t$ has the same size as the coarse flow $\hat{F}_{coarse}^t$, adding pixel-wise offsets to $\hat{F}_{coarse}^t$ to precisely align with the generated layout $\hat{LO}_C^t$. Thus our final transformation flow is given by: $\hat{F}^t=\hat{F}_{coarse}^t+\hat{F}_{fine}^t$. Using $\hat{F}^t$ to warp the exemplary clothes ${FG}_{C}$, we can obtain the final warped clothes $\hat{FG}_C^t$.\par
During training, we also utilize a VGG loss \shortcite{johnson2016perceptual} $L_{VGG}$ to minimize the difference between $\hat{FG}_C^t$ and the ground truth, which enables the supervision of the fine warping.\par
\subsection{Layout-Constrained Deformable Convolution}
Since both the coarse and the refinement flows are predicted based on FPN-A and FPN-M features, feature extraction in these two FPNs directly affects the quality of our warping results. In motion transfer tasks, clothes items may change to various shapes along with body poses. Thus the extracted features should be able to generalize to various shapes correspondingly. Unfortunately, standard CNN features are transformation-invariant, which means receptive fields remain fixed no matter how the shape changes and hence can't accommodate the geometric deformations for different shapes.
Besides, such fixed receptive fields are not large enough to accommodate the misalignment between appearance and motion features.\par
\begin{figure}[t]
\centering
\includegraphics[width=0.7\columnwidth]{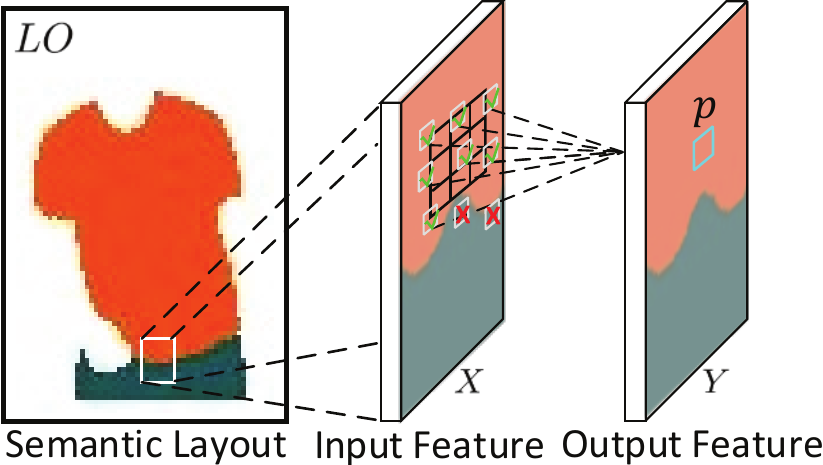} 
\caption{Illustration of LC-DConv. Here we take the LC-DConv in the first layer of FPN-M as an example, which is the same for FPN-A expect FPN-A takes foregrounds in addition to layouts as its inputs. We let orange and dark green represent semantic classes of tops and bottoms, respectively. In the small patch, $\checkmark$ in green and $\times$ in red denote valid and invalid sampling positions, respectively.}
\label{DConv}
\end{figure}
Therefore, we replace all the standard convolutions in bottom-up pathway layers of FPN-A and FPN-M with Deformable Convolutions (DConv) \cite{dai2017deformable}, which can model geometric deformations adaptively with deformable receptive fields. As shown in Figure \ref{DConv}, DConv learns additional 2D offsets to shift regular sampling locations of the standard convolution, which enables deformable and larger receptive fields. However, the unconstrained offsets may result in invalid sampling from positions not semantically related to the output position, causing the loss of semantic information in the output feature. Therefore, our Layout-Constrained Deformable Convolution (LC-DConv) utilizes input semantic layouts as priors to set amplitudes of features sampled from invalid positions to zero, precisely preserving the layout boundaries and thus enhancing the semantic information in the output feature.
Taking the convolution with a 3x3 kernel of dilation 1 as an example, we explain how our LC-DConv works. Let $X(p)$ and $Y(p)$ be the input and the output features at position $p$ respectively, and let $w_k$, $\Delta{p_k}$ and $p_k{\in}\{(-1,-1),(-1,0),...,(1,1)\},k=1{\sim}9$ represent the $k$-th kernel weight, the $k$-th sampling offset and the $k$-th regular sampling position respectively, we can derive the LC-DConv as follows:
\begin{equation}
\begin{aligned}
Y(p)=&\sum_{k=1}^K{w_k\cdot{X(p+p_k+\Delta{p_k})\cdot\Delta{m_k}}},\\
\Delta{m_k}=&
\begin{cases}
0,& \text{$LO(p)\,\,{\neq}\,LO(p+p_k+\Delta{p_k})$},\\
1,& \text{otherwise,}
\end{cases}
\label{con:E1}
\end{aligned}
\end{equation}where $K=9$, $\Delta{m_k}$ is the modulation scalar determined by the layout prior $LO$, deciding the validity of the $k$-th offset sampling position. For FPN-A, $LO$ refers to the exemplary clothing layout ${LO}_{C}$. For FPN-M, $LO$ refers to the generated clothing layout $\hat{LO}_C^t$. As depicted in Figure \ref{DConv}, we set feature amplitudes to zero if they belong to semantic classes different from the class at the output position, which can effectively avoid any invalid sampling.\par
\begin{figure}[t]
\centering
\includegraphics[width=0.7\columnwidth]{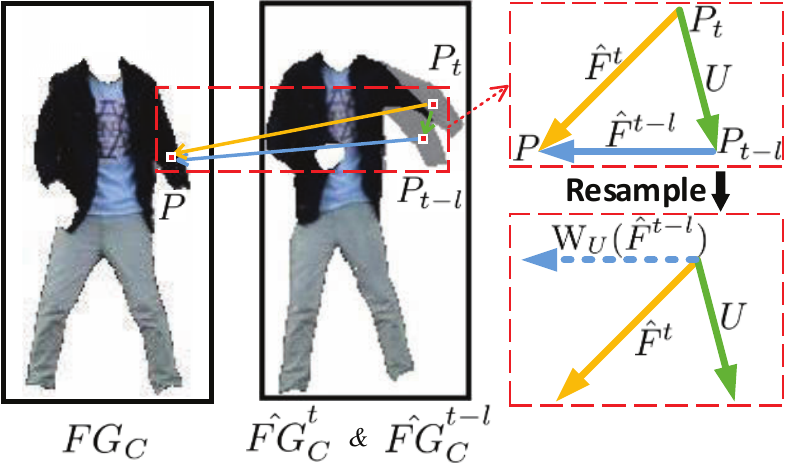} 
\caption{Illustration of FTC loss. The first image is the exemplary clothes ${FG}_{C}$. The second image is a combination of two warped clothes $\hat{FG}_C^{t-l}$ and $\hat{FG}_C^t$. We take the position $P$ in the right arm region as an example, which moves from $P_{t-l}$ to $P_t$ during time $t-l{\sim}t$. Yellow, blue and green solid arrows represent $\hat{F}^{t}$, $\hat{F}^{t-1}$ and $U$, respectively. Blue dotted arrow denotes the resampled $\hat{F}^{t-1}$.}
\label{FTC}
\end{figure}
\subsection{Flow Temporal Consistency Loss}
Compared to other methods \cite{chan2019everybody,wang2019few} that only learn implicit temporal consistency among frame images, our FTC loss uses optical flows to enforce explicit temporal consistency among transformation flows, also enabling direct supervision on transformation flows instead of warped clothes. Specifically, benefiting from the flow format of our transformation, we can build the correlation between two transformation flows $\hat{F}^t$ and $\hat{F}^{t-l}$ using the optical flow $U$ between the corresponding two frames. Let $P=(x,y)$ denote the position of a pixel in the exemplary clothes ${FG}_{C}$, and let $P_t=(x_t,y_t)$ and $P_{t-l}=(x_{t-l},y_{t-l})$ denote positions of the same pixel in the warped clothes $\hat{FG}_C^t$ and $\hat{FG}_C^{t-l}$. Thus, for this pixel, the transformation flow vectors at time $t$ and $t-l$ are given by:
\begin{equation}
\begin{aligned}
&\hat{F}^t(x_t,y_t)=(x-x_t,y-y_t),\\
&\hat{F}^{t-l}(x_{t-l},y_{t-l})=(x-x_{t-l},y-y_{t-l}).
\label{con:E2}
\end{aligned}
\end{equation}Note that all the flows used in this paper are backward flows. Therefore, the flow vectors at time steps $t$ and $t-l$ are actually located at the transformed positions $P_t$ and $P_{t-l}$ w.r.t. the warped clothes, rather than the original position $P$ w.r.t. the exemplary clothes. Such backward format can ensure each pixel in the warped clothes has a flow vector to indicate its original position to be sampled from the exemplary clothes, further ensuring the warping operation is valid.\par
In principle, if the frames at $t$ and $t-l$ are temporally consistent, $\hat{F}^t(x_t,y_t)-\hat{F}^{t-l}(x_{t-l},y_{t-l})$ should be equal to the ground-truth optical flow vector $U(x_t,y_t)$, which is from $t$ to $t-l$ and equal to $(x_{t-l}-x_t,y_{t-l}-y_t)$. As shown in Figure \ref{FTC}, to generalize this equation to the whole image rather than a single pixel, we should resample $\hat{F}^{t-l}(x_{t-l},y_{t-l})$ at $P_{t-l}$ to the same position $P_t$ as $\hat{F}^t(x_t,y_t)$, and do the same to all the remaining pixels of $\hat{F}^{t-l}$ to make them share positions with those of $\hat{F}^t$. We can realize this by using $U$ to warp $\hat{F}^{t-l}$. Thus our FTC loss is given by:
\begin{equation}
\begin{aligned}
L_{FTC,l}=\|\hat{F}^t-\mathrm{W}_{U}(\hat{F}^{t-l})-{U}\|_1,
\label{con:E3}
\end{aligned}
\end{equation}where $\mathrm{W}$ denotes the warping operation based on $U$. To guarantee both short-term and long-term temporal consistency, we set $l=1,3,9$ to compute FTC losses at three time scales and sum them together as our full FTC loss $L_{FTC}$.\par
We further utilize a TVL1 loss \cite{fan2018end} $L_{TVL1}$ to minimize the difference between flow vectors at neighbored positions of $\hat{F}^t$, which smooths the warping.
Summarily, the full objective is a weighted sum of several losses, given by:
\begin{equation}
\begin{aligned}
L_{full}=L_{VGG}+L^{coarse}_{VGG}+\lambda_{1}L_{FTC}+\lambda_{2}L_{TVL1},
\label{con:E4}
\end{aligned}
\end{equation}where $\lambda_{1}$ and $\lambda_{2}$ denote the weights of FTC and TVL1 losses, respectively.
\begin{table*}[htbp]
\begin{center}
\caption{Quantitative results tested on our SoloDance dataset and iPER \shortcite{liu2019liquid} dataset. SSIM, PSNR, TCM are similarity metrics, the higher the better (SSIM and TCM range from 0 to 1). LPIPS and FID are distance metrics, the lower the better. Note that TCM measures temporal consistency while other metrics measure spatial consistency.}\label{table1}
\resizebox{1\textwidth}{!}{
\begin{tabular}{c"lccccc"cc"c}
  \thickhline
  Datasets & Metrics & EDN\shortcite{chan2019everybody} & FSV2V\shortcite{wang2019few} & LWGAN\shortcite{liu2019liquid} & SGWGAN\shortcite{dong2018soft} & ClothFlow\shortcite{han2019clothflow} & w/o FTC loss & w/o LC-DConv & Ours\\
  \thickhline 
  & SSIM & 0.811 & 0.721 & 0.786 & 0.763 & 0.843 & 0.849 & 0.850 & \textbf{0.879}\\
  & PSNR & 23.22 & 20.84 & 20.87 & 20.54 & 22.06 & 23.05 & 23.19 & \textbf{26.65}\\
  SoloDance & LPIPS & 0.051 & 0.132 & 0.106 & 0.124 & 0.072 & 0.065 & 0.063 & \textbf{0.049}\\
  & FID & 53.17 & 112.99 & 86.53 & 99.24 & 76.61 & 64.92 & 61.03 & \textbf{46.49}\\
  & TCM & 0.347 & 0.106 & 0.176 & 0.166 & 0.322 & 0.319 & 0.401 & \textbf{0.641}\\
  \thickhline 
  & SSIM & 0.840 & 0.780 & 0.825 & 0.818 & 0.814 & 0.824 & 0.822 & \textbf{0.849}\\
  & PSNR & 23.39 & 20.44 & 21.43 & 22.41 & 21.87 & 22.76 & 22.52 & \textbf{24.27}\\
  iPER\shortcite{liu2019liquid} & LPIPS & 0.076 & 0.110 & 0.091 & 0.086 & 0.088 & 0.082 & 0.084 & \textbf{0.072}\\
  & FID & 56.29 & 110.99 & 77.99 & 101.99 & 71.21 & 64.40 & 63.72 & \textbf{55.07}\\
  & TCM & 0.361 & 0.184 & 0.197 & 0.260 & 0.422 & 0.411 & 0.499 & \textbf{0.687}\\
  \thickhline
\end{tabular}}
\end{center}
\end{table*}
\begin{figure}[t]
\centering
\includegraphics[width=1\columnwidth]{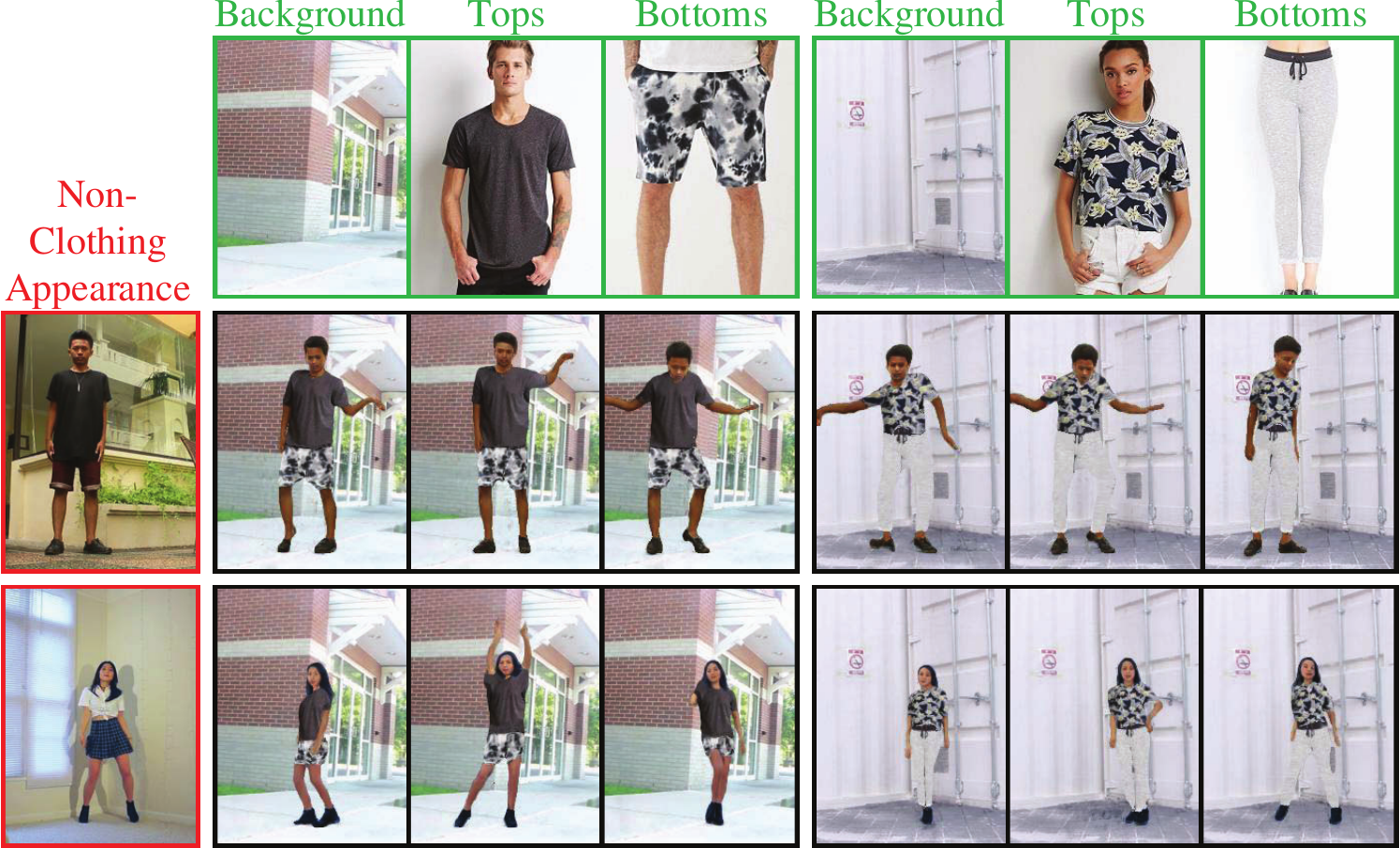} 
\caption{Examples of our multi-source appearance synthesis. Red-edged images describe the non-clothing foreground appearances (hair, face, torso, shoes). Green-edged images from left to right describe the appearances of background, tops, bottoms. Black-edged images are our synthesized motion transfer results. \emph{Please zoom in for a better view.}}
\label{appearance}
\end{figure}
\subsection{Multi-Source Appearance Attribute Editing}
Compared with existing HVMT methods, C2F-FWN can support multi-source appearance attribute editing when transferring motions.
As described above, we divide the exemplary appearance into the background, clothing and non-clothing foregrounds. The clothing foreground can be further divided into tops and bottoms, which decide how the exemplary human subject is dressed in the synthesized videos. The non-clothing foreground can be further divided into hair, face, torso and shoes, with the first three parts deciding the human identity.
With the help of semantic layouts, the background and each part of the foregrounds can be extracted from different sources to achieve the multi-source exemplary appearance. For example, the background can be replaced by arbitrary fixed images. Tops and bottoms in the clothing foreground can be extracted from arbitrary fashion or portrait images, which is the same for parts of the non-clothing foreground.
With such multi-source appearance inputs, our proposed method can generate the corresponding multi-source appearance in the synthesized videos, which enables full appearance attribute editing for motion transfer as shown in Figure \ref{appearance}.
Such capability can achieve rather high flexibility and efficiency in real applications.
For example, users can arbitrarily change their clothes and backgrounds in videos without really wearing these clothes or performing in these backgrounds, enabling convenient video re-creation.
\section{Experiments}
\subsection{Dataset}
\subsubsection{SoloDance Dataset}
We built a large-scale SoloDance dataset containing 179 solo dance videos with 53,700 frames.
Specifically, 143 human subjects were captured with each wearing various clothes and performing complex dances (e.g., modern, street dances) in various backgrounds. 
Compared to the iPER dataset \cite{liu2019liquid} that only contains 30 subjects performing simple moves (e.g., random actions, A-poses), our dataset offers more appearance variety and motion complexity.
We utilized \cite{cao2017realtime} and \cite{gong2018instance} to detect body poses and semantic layouts, and further obtained foregrounds and backgrounds for each video.
In our experiments, we randomly split the dataset into 153 and 26 videos for training and testing.
\subsubsection{iPER Dataset}
We also evaluated our method on the iPER dataset \shortcite{liu2019liquid}. The data preprocessing of the iPER dataset is the same as our SoloDance dataset. Following the original protocal of iPER, we used 164 videos for training and the remaining 42 videos for testing.
\subsection{Implementation Details}
All the frames were resized and cropped to 256x256 sizes to train our models. Since backgrounds are fixed and easy to generate compared to animated human foregrounds, we further cropped the frames to central 192x256 body regions during evaluation to focus on the quality of the synthesized foregrounds.
The design of the layout GAN in Stage 1 and the composition GAN in Stage 3 followed \cite{wang2018video}.
The design of our FPNs in Stage 2 followed \cite{lin2017feature} except that we replaced standard convolutions in bottom-up pathways of FPN-A and FPN-M with our LC-DConv to enhance the features. Particularly, the LC-DConv was implemented based on \cite{dai2017deformable} by employing layout-constrained sampling locations. Moreover, to enable the supervision of the proposed FTC loss, we utilized \cite{ilg2017flownet} to obtain the ground-truth optical flows.
We trained each stage for 10 epochs separately with Adam optimizers \cite{kingma2014adam} (learning rate: 0.0002, ${\beta}_1$: 0.5, ${\beta}_2$: 0.999) on an Nvidia RTX 2080 Ti GPU, where we set $\lambda_{1}=5$ and $\lambda_{2}=0.5$ in Eq. \ref{con:E4} to trade-off the two losses.
\subsection{Baselines}
To evaluate our proposed approach, we made comparisons with state-of-art HVMT methods including a personalized method EDN \cite{chan2019everybody}, a direct generation method FSV2V \cite{wang2019few}, two feature warping methods LWGAN \cite{liu2019liquid} and SGWGAN \cite{dong2018soft}, and an image warping method ClothFlow \cite{han2019clothflow}.
In our implementation, we used 3000 frames for each person to train personalized models for EDN, and used the same data as ours to train models for other methods.
\begin{figure*}[htbp]
\centering
\includegraphics[width=.95\textwidth]{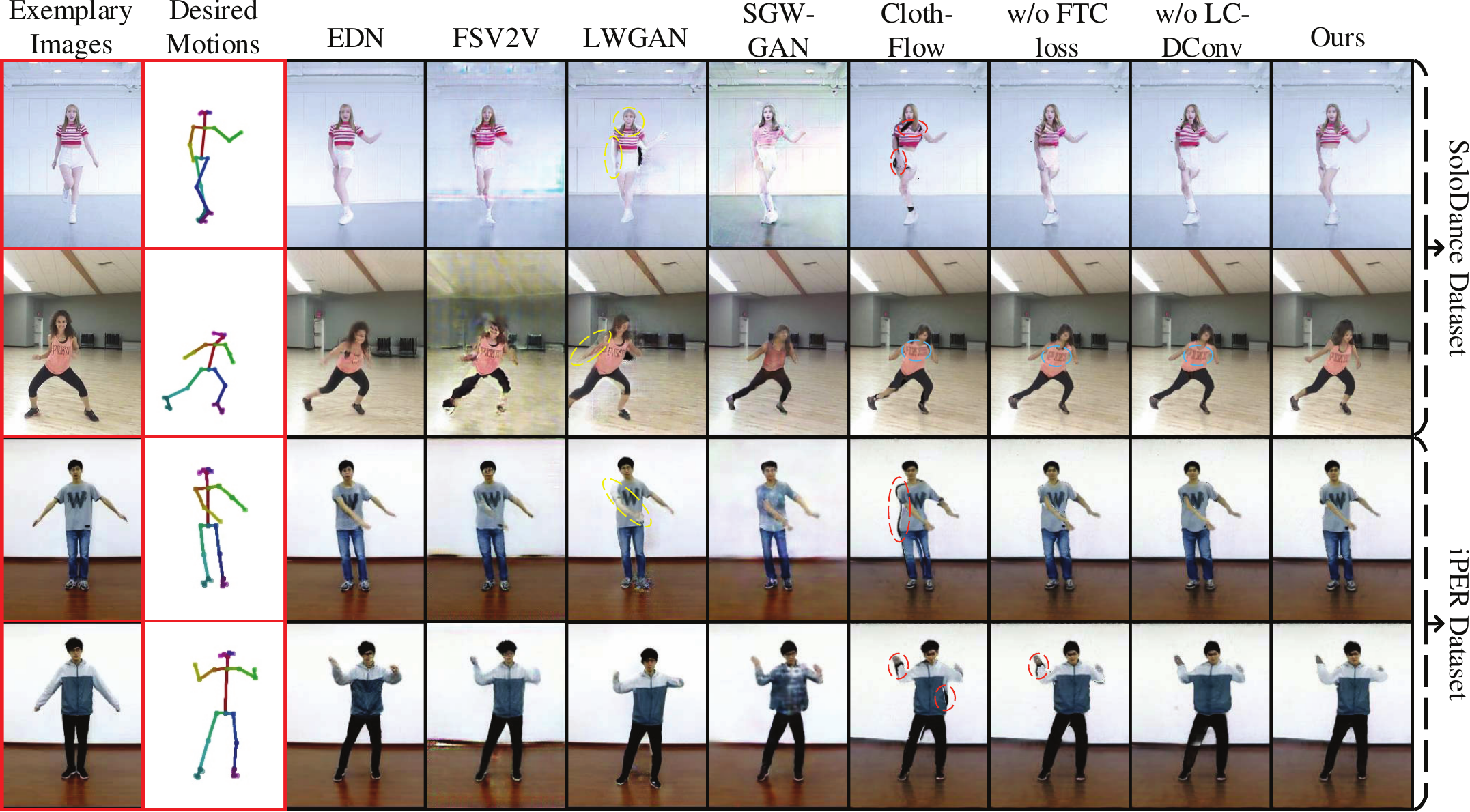} 
\caption{Qualitative comparisons with other methods including EDN \shortcite{chan2019everybody}, FSV2V \shortcite{wang2019few}, LWGAN \shortcite{liu2019liquid}, SGWGAN \shortcite{dong2018soft}, ClothFlow \shortcite{han2019clothflow}, albated variants without FTC loss, LC-DConv. Yellow, blue and red circles point out blurry surfaces, over-stretched clothes patterns, and black chinks (caused by misplacements), respectively. \emph{Please zoom in for a better view.}}
\label{comparison}
\end{figure*}
\subsection{Quantitative Results}
We utilized both traditional (SSIM and PSNR) and CNN-based metrics (LPIPS \cite{zhang2018unreasonable} and FID \cite{heusel2017gans}) to measure the quality of synthesized frames, which can assess the spatial consistency between synthesized and exemplary images.
We also utilized a Temporal Consistency Metric (TCM) \cite{yao2017occlusion} to evaluate the temporal consistency, which is an essential factor in measuring the quality of videos rather than single frames. Specifically, TCM measures temporal consistency by calculating warping errors between successive synthesized frames, where each frame is warped by the ground-truth optical flow to compare with its neighbored frame.
The quantitative results of all the methods are summarized in Table \ref{table1}. We can see that our proposed C2F-FWN significantly outperforms all the other methods including the personalized method EDN \shortcite{chan2019everybody} for all the metrics (especially for the TCM scores) on both of the two datasets, which indicates that our approach can achieve HVMT with better spatial and temporal consistency.
\subsection{Qualitative Results}
As shown in Figure \ref{comparison}, we randomly visualize some motion transfer video frames synthesized by different methods for qualitative comparisons, where our approach outperforms all the other methods.
Specifically, we achieve better spatial consistency with exemplary images than others, especially the direct generation method FSV2V \shortcite{wang2019few}, where our method preserves the exemplary appearance details such as decorative patterns and colors well.
Besides, benefiting from our coarse-to-fine flow warping, we can capture the desired motions better than existing warping-based methods, with our warped clothes precisely aligned with the body layouts. However, the feature warping method SGWGAN \shortcite{dong2018soft} can't enable precise feature alignment with the desired motions due to the limited warping capability, which causes poor appearance details.
Another feature warping method LWGAN \shortcite{liu2019liquid} results in blurry details on the surface of bodies and clothes (e.g., circled in yellow in Figure \ref{comparison}) because of the low-precision SMPL models.
The image warping method ClothFlow \shortcite{han2019clothflow} can't warp the exemplary images to align with the desired motions, which results in visual artifacts such as over-stretching and misplacement near the layout boundaries (e.g., circled in blue and red in Figure \ref{comparison}).
Although the personalized method EDN \shortcite{chan2019everybody} can generate comparable results to us, it often results in blurrier textures.
We also show some of our multi-source appearance synthesis results in Figure \ref{appearance}, where we utilized fashion images dissimilar from our training data to extract tops and bottoms. We can see that the multi-source exemplary appearances are also well preserved, enabling flexible appearance attribute editing for HVMT.
Videos of the qualitative comparisons and our synthesized results can be found in \emph{our supplementary materials}, where we show that our method can also achieve better temporal consistency.
\subsection{Ablation Study}
We also conducted ablation studies w.r.t. our FTC loss and LC-DConv to demonstrate their effectiveness.
Specifically, we implemented two variant models for comparisons. One was trained without our FTC loss, and another only adopted standard convolutions to extract features.
As shown in Table \ref{table1}, our full method outperforms the two variants for all the metrics.
As shown in Figure \ref{comparison}, without the two components, the clothes are warped imprecisely (e.g., circled in blue and red in Figure \ref{comparison}), which indicates the importance of the LC-DConv as well as the FTC loss for enhancing our flow warping and improving spatial consistency.
Moreover, we observed that the variant without the FTC loss would result in much worse video coherence than our full method, which shows our superiority in improving temporal consistency. \textbf{Please refer to \emph{our supplementary video} for more details: \url{https://youtu.be/THuQN1GXuGI}.}
\section{Conclusion}
In this paper, we propose Coarse-to-Fine Flow Warping Network (C2F-FWN) to achieve both spatial and temporal consistency for HVMT, enabling us to preserve exemplary appearances as well as improve video coherence. Specifically, our coarse-to-fine flow warping can precisely model geometric deformations caused by motions to ensure the spatial consistency, where we further utilize our Layout-Constrained Deformable Convolution (LC-DConv) to enhance features for estimating the transformation flows.
To achieve the temporal consistency, we propose a novel Flow Temporal Consistency (FTC) Loss to learn explicit temporal consistency between successive transformation flows, which significantly improves the video coherence.
Experimental results tested on our SoloDance dataset and the iPER dataset show our superiority to other methods in terms of both spatial and temporal consistency. Ablation studies w.r.t. our FTC loss and LC-DConv demonstrate their effectiveness in improving our synthesis quality. We also demonstrate that our method can achieve flexible appearance attribute editing provided with alterable multi-source appearance inputs, which shows promising application prospects.
\subsection{Limitations and Future Work}
Although our method works well in most cases, it may fail (e.g., jitters, blurs) due to errors in poses and semantic layouts, which would cause errors in our model inputs and further result in artifacts in our output results. In the future, we can utilize more accurate pose and layout estimation techniques to eliminate these errors.
Besides, we currently only provide our model with one single exemplary image, which might suffer from self occlusions and texture missings in cases of extremely large motion changes. Thus, how to attend and aggregate multiple exemplary images for warping is also worth studying in future works.

\section{Acknowledgments}
This work was supported by the National Natural Science Foundation of China (U19B2043).
\bibliography{bibfile.bib}
\end{document}